\documentclass[fleqn,10pt]{wlscirep}
\usepackage[utf8]{inputenc}
\usepackage[T1]{fontenc}
\usepackage{tabularx}
\usepackage{booktabs}
\usepackage{caption}
\usepackage{url}
\usepackage{lineno}
\usepackage{multirow}
\usepackage{amsmath}
\usepackage{makecell}
\usepackage[normalem]{ulem}
\usepackage{fontawesome5}
\usepackage{tikz}
\usetikzlibrary{positioning, shapes.multipart, arrows.meta, fit, backgrounds}
\usetikzlibrary{positioning,shapes,fit}
\usepackage{fontawesome5}
\usepackage{enumitem}
\usepackage{tabularx}
\usepackage{xcolor}
\usepackage{titlesec}
\usetikzlibrary{positioning, arrows.meta, shapes.multipart, decorations.pathreplacing}
\usetikzlibrary{shapes, backgrounds}
\usetikzlibrary{calc}

\title{PERCS: Persona-Guided Controllable Biomedical Summarization Dataset}

\author[1,*]{Rohan Charudatt Salvi}
\author[2]{Chirag Chawla}
\author[3]{Dhruv Jain}
\author[3]{Swapnil Panigrahi}
\author[3]{Md Shad Akhtar}
\author[1]{Shweta Yadav}
\affil[1]{Department of Computer Science, University of Illinois, Chicago, IL, 60607, USA}
\affil[2]{Indian Institute of Technology, Varanasi, India}
\affil[3]{Department of Computer Science \& Engineering, Indraprastha Institute of Information Technology, Delhi, New Delhi, 110020, India}

\affil[*]{corresponding author email: rcsalvi2@uic.edu}

\begin{abstract}
Automatic medical text simplification plays a key role in improving health literacy by making complex biomedical research accessible to diverse readers. However, most existing resources assume a single generic audience, overlooking the wide variation in medical literacy and information needs across user groups. To address this limitation, we introduce \textit{PERCS} (\underline{PER}sona-guided \underline{C}ontrollable \underline{S}ummarization), a dataset of biomedical abstracts paired with summaries tailored to four personas: \textit{Laypersons}, \textit{Premedical Students}, \textit{Non-medical Researchers}, and \textit{Medical Experts}. These personas represent different levels of medical literacy and information needs, emphasizing the need for targeted, audience-specific summarization. Each summary in PERCS was reviewed by physicians for factual accuracy and persona alignment using a detailed error taxonomy. Technical validation shows clear differences in readability, vocabulary, and content depth across personas. Along with describing the dataset, we benchmark four large language models on PERCS using automatic evaluation metrics that assess comprehensiveness, readability, and faithfulness, establishing baseline results for future research. The dataset, annotation guidelines, and evaluation materials are publicly available to support research on persona-specific communication and controllable biomedical summarization.
\end{abstract}
\begin{document}

\flushbottom
\maketitle
%
%

\section*{Background \& Summary}
The internet has become a primary source of health information for the public. Surveys indicate that nearly four in five internet users in the United States seek medical or health-related information online \cite{fox2013health}, and an increasing proportion now rely on responses generated by large language models (LLMs) to address their healthcare questions \cite{adams_2025}. While such models can produce fluent and informative text, they are also prone to factual inaccuracies and hallucinations \cite{pal-etal-2023-med, yadav2023towards,yadav2021reinforcement}. Scientific repositories such as PubMed \cite{national_library_of_medicine_2025} provide a means to verify these claims, yet the biomedical literature they contain is often dense with specialized terminology and complex sentence structures that make it difficult for non-expert readers to comprehend and apply \cite{korsch1968gaps, friedman2002two}. This gap between scientific communication and public understanding has long been recognized, motivating the development of plain-language summaries, which translate complex biomedical findings into language that is accurate, accessible, and understandable to diverse audiences \cite{stableford2007plain}. Plain-language summarization aims to simplify scientific text while maintaining its essential meaning, but current automatic approaches generally adopt a “one size fits all” assumption about a single “non-expert” audience. In reality, individuals differ widely in their educational backgrounds, familiarity with biomedical concepts, and information needs \cite{tran-etal-2025-readctrl, mullick-etal-2024-persona}. As a result, a summary appropriate for a layperson may not meet the expectations of a pre-medical student, an interdisciplinary researcher, or a medical professional. Recent advances in controllable and readability-based summarization have introduced mechanisms for adjusting lexical complexity and sentence structure \cite{luo-etal-2022-readability, zhang-etal-2024-atlas}, yet these systems primarily emphasize readability rather than content personalization \cite{tran-etal-2025-readctrl}. Furthermore, the datasets required to develop and evaluate persona-specific biomedical summaries remain limited.

Existing corpora have advanced plain-language summarization but offer limited support for audience-specific adaptation. Early datasets such as CDRS \cite{guo2021automated}, PLOS, and eLife \cite{goldsack-etal-2022-making}, PLABA \cite{attal2023dataset} pair biomedical research articles with plain-language summaries written for a general public audience, focusing primarily on improving readability rather than persona adaptation. More recent efforts have introduced elements of persona-based summarization. 
The WebMD-derived corpus \cite{mullick-etal-2024-persona} categorizes readers into three discrete groups, namely doctor, patient, and general reader, enabling the creation of fine-grained, persona-specific summaries. Similarly, MedEasi \cite{basu2023med} includes two personas, expert and layperson, and provides corresponding summaries of biomedical texts along with edit-level annotations that capture simplification operations such as elaboration, deletion, and replacement. These resources represent important progress toward personalized biomedical communication but remain limited in scope and granularity. They typically offer only one or two persona types per document and cover a narrow range of medical topics. Consequently, there is still a need for a larger, systematically curated dataset that provides multiple parallel summaries reflecting distinct levels of reader expertise for the same biomedical source material.

To address this gap, we present the PERCS (\underline{PER}sona-guided \underline{C}ontrollable \underline{S}ummarization) dataset, an expert-annotated corpus designed to facilitate research on audience-aware summarization in the biomedical domain. PERCS comprises 500 biomedical research abstracts drawn from PubMed, each paired with four distinct summaries written for readers with different expertise levels: laypersons, pre-medical students, researchers from non-medical fields, and medical experts. This structure yields 2,000 persona-specific summaries that vary systematically in terminology, information depth, and readability. As a result, PERCS summaries differ in structure, tone, informational depth, and style, as illustrated in Figure \ref{fig:persona-summary-examples}. Lay summaries emphasize clarity and minimize jargon, describing CRP as “\textit{a protein used to check for bacterial infection.}” Pre-med summaries introduce basic terminology, defining CRP as “\textit{a substance produced by the liver in response to inflammation.}” Researcher summaries focus on methods and findings, describing CRP as “\textit{a marker to detect infections},” while expert summaries retain domain-specific terminology and quantitative details. Each summary was created and reviewed by medical experts following detailed annotation guidelines to ensure factual consistency and appropriateness for the intended audience. By aligning multiple versions of the same text to clearly defined reader personas, PERCS provides an expert-curated resource for studying how biomedical information can be tailored to different personas. The dataset serves as a reference for developing and evaluating systems that create personalized biomedical summaries suited to readers with different backgrounds and levels of understanding. To guide future research, we also evaluate several state-of-the-art language models on PERCS across multiple aspects, such as comprehensiveness, readability, and factuality. Together, these contributions lay the groundwork for building and benchmarking models that produce accurate, persona-aware summaries of biomedical information.

\begin{figure*}[t]
\centering
\begin{tikzpicture}[
    box/.style={rectangle, draw=blue!70!black, rounded corners, text width=4.2 cm,
                align=left, inner sep=3pt, font=\footnotesize},
    example/.style={rectangle, draw=blue!70!black, rounded corners, text width=12.7cm,
                align=left, inner sep=3pt, font=\footnotesize}
]

\node[box, minimum height=1.4 cm] (abstract) {
    \textbf{Persona}\\
};

\node[box, below=0.1cm of abstract, minimum height=1.5 cm] (lay) {
    \textbf{Layperson}\\
    • Avoid medical jargon.\\
    • Overview of purpose, method, results.\\
    • Use short, simple sentences.
};

\node[box, below=0.12cm of lay] (premed) {
    \textbf{Pre-med}\\
    • Briefly explain medical terms.\\
    • Clear, structured sentences.\\
    • Emphasize method and findings.
};

\node[box, below=0.1cm of premed, minimum height=1.75 cm] (researcher) {
    \textbf{Researcher}\\
    • Accessible scientific language.\\
    • Clear and formal sentence. \\
    • Detailed design, method, and results.
};

\node[box, below=0.15cm of researcher] (expert) {
    \textbf{Expert}\\
    • Field-specific terminology.\\
    • Concise, professional tone.\\
    • Quantitative detail, design, findings.
};

\node[example, right=0.2cm of abstract] (abstract_ex) {
    \textbf{Abstract:} Purpose: To evaluate the practical value of initial C-reactive protein (CRP) in the diagnosis of bacterial infection in children. Methods: The subjects comprised 11 children, six boys and five girls, aged 3 months through to 3 years (median age 16 months), whose initial CRP levels were < 1.0 mg/dL despite bacterial infection....
};

\node[example, right=0.2cm of lay, yshift=-0.02cm] (lay_ex) {
    \textbf{Summary:} The purpose of this study was to see how useful a 
    \textcolor{green!50!black}{\uline{certain protein, called C-reactive protein (CRP), is in figuring out if a child has a bacterial infection}}. 
    \textcolor{green!50!black}{\uline{This protein usually increases in the body when there's an infection}}, so doctors often measure it to help diagnose illnesses. 
    The study looked at {\uline{11 children, a mix of boys and girls, who were between 3 months and 3 years old}}. 
    Even though these kids had bacterial infections, their  \textcolor{green!50!black}{\uline{initial CRP levels were surprisingly low}}...
};

\node[example, right=0.2cm of premed, yshift=-0.01cm] (premed_ex) {
    \textbf{Summary:} The purpose of this study was to assess how useful the initial levels of 
    \textcolor{green!50!black}{\uline{C-reactive protein (CRP), a substance produced by the liver in response to inflammation}}, are in diagnosing bacterial infections in children. 
    The study involved {\uline{11 children (six boys and five girls, aged 3 months to 3 years)}}, who had bacterial infections but initially showed 
    \textcolor{blue!70!black}{\uline{CRP levels below 1.0 mg/dL, which is considered low}}...
};

\node[example, right=0.2cm of researcher, yshift=-0.05cm] (researcher_ex) {
     \textbf{Summary:} The study aimed to assess the effectiveness of using initial C-reactive protein (CRP) levels in diagnosing bacterial infections in children. 
     \textcolor{green!50!black}{\uline{CRP is a protein produced by the liver in response to inflammation, often used as a marker to detect infections or diseases}}. 
     The study involved \textcolor{blue!70!black}{\uline{11 children (six boys and five girls, aged 3 months to 3 years)}}, who had bacterial infections but initially showed 
     \textcolor{blue!70!black}{\uline{low CRP levels (< 1.0 mg/dL)}}...
};

\node[example, right=0.2cm of expert, yshift=-0.005cm, minimum height=1.8 cm] (expert_ex) {
    \textbf{Summary:} This study assessed the {\uline{diagnostic value of initial C-reactive protein (CRP) levels in detecting bacterial infections in children}}. The study involved \textcolor{blue!70!black}{\uline{11 children (six boys and five girls) aged 3 months to 3 years (median age: 16 months)}}, who initially presented with 
    \textcolor{blue!70!black}{\uline{CRP levels below 1.0 mg/dL despite having bacterial infections}}...
};

\end{tikzpicture}

\caption[Persona-specific summarization of a biomedical abstract in PERCS.]%
{Persona-specific summarization of a biomedical abstract in PERCS. 
{\textcolor{green!50!black}{Green}} represents {\textcolor{green!50!black}{simplification}}, and 
{\textcolor{blue!70!black}{Blue}} represents {\textcolor{blue!70!black}{information detail}}.}

\label{fig:persona-summary-examples}
\end{figure*}

\vspace{-1em}

\begin{table*}
\small
\centering
\caption{Different error types along with their description and examples in the persona-aware biomedical summarization.}
\label{tab:error-types}
\begin{tabularx}{\textwidth}{p{0.17\textwidth} p{0.4\textwidth} p{0.38\textwidth}}
\toprule
\textbf{Error} & \textbf{Description} & \textbf{Example} \\
\midrule
Incorrect Definitions & Wrong or misleading explanations of medical terms or concepts. & Defining diabetes or insulin function inaccurately. \\

Incorrect Synonyms & Replacing medical words with inaccurate or oversimplified terms. & Using “painkillers” for anti-inflammatory drugs. \\

Incorrect Background & False or irrelevant contextual details about prevalence or treatment. & Calling a rare disorder common or misdescribing therapy. \\

Entity Errors & Wrong factual details like numbers, names, or dosages. & Reporting “50 mg” instead of “5 mg” or “80” vs. “800” patients. \\

Contradiction & Summary directly opposes the abstract’s results or claims. & Abstract: “Drug A reduced symptoms.” Summary: “No effect.” \\

Omission & Missing key findings or results from the abstract. & Skipping main outcomes or notable side effects. \\

Jumping to Conclusions & Overstating results beyond what data supports. & Abstract: “may help,” but summary: “proven cure.” \\

Misinterpretation & Misstating or oversimplifying meaning of the abstract. & Saying “FDA approved” when text says “FDA allowed.” \\

Structure Error & Disorganized layout or mixing sections like methods and results. & Writing results under background or as FAQ style. \\

Persona Relevance & Language complexity unsuitable for target persona. & Using jargon for lay readers or oversimplifying expert text. \\

Hallucination & Adding fabricated or irrelevant content not in abstract. & Mentioning an unrelated drug or disease. \\
\bottomrule
\end{tabularx}
\end{table*}

\section*{Methods}

\subsection*{Problem Statement}

To support the creation of persona-aware biomedical summaries, we define four reader personas representing distinct levels of expertise and information needs. Each biomedical abstract in the PERCS dataset is paired with four summaries, one for each persona:

\begin{enumerate}
\item \textit{\textbf{Layperson:}} Readers with a high school–level understanding of biology and limited medical knowledge. Summaries use simple language, avoid jargon, and emphasize the study’s purpose and main findings.

\item \textit{\textbf{Pre-med:}} Students in the early stages of health-related higher education who are familiar with basic medical terminology. Summaries include concise definitions of key terms and describe methods and results in clear, structured sentences.

\item \textit{\textbf{Researchers:}} Scientifically trained individuals without biomedical specialization, such as engineers, policymakers, or clinicians from other fields. Summaries describe study design and key results using accessible scientific language.

\item \textit{\textbf{Medical experts:}} Biomedical researchers and practitioners. Summaries remain concise and technical, retaining domain-specific terminology, methodological details, quantitative results, and findings.
\end{enumerate}

\begin{figure*}
    \centering
    \includegraphics[width=0.85\linewidth]{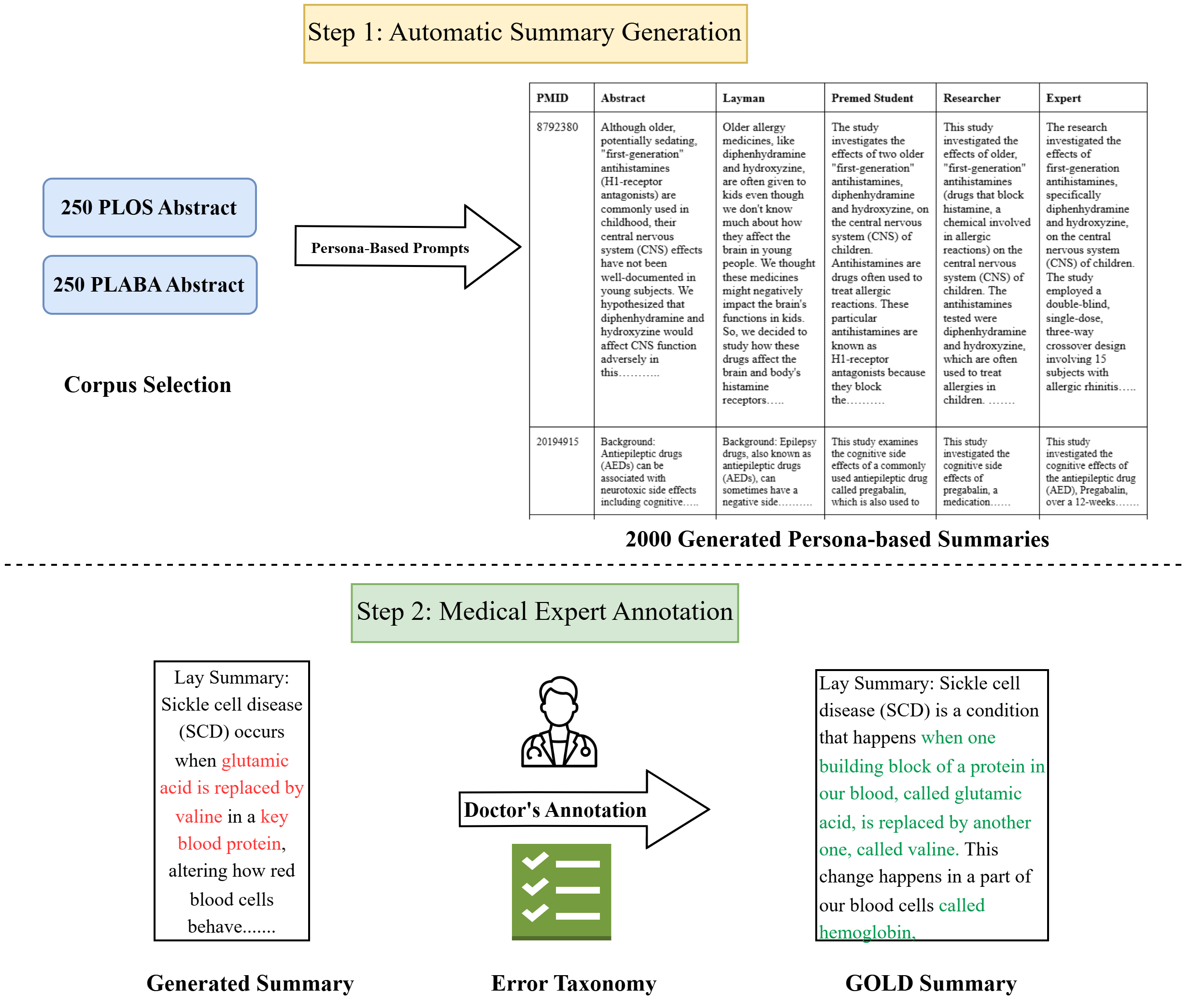}
    \caption{An overview of the steps involved in constructing the PERCS dataset, namely data collection, summary generation, and expert validation.}
    \label{fig:PERCS_overview_hor}
\end{figure*}

These four personas represent the major audience groups commonly addressed in science communication: the general public with limited medical literacy \cite{national2017communicating}, learners progressing toward medical expertise, adjacent but non-specialist researchers \cite{van2024understanding}, and domain experts who require technical precision. Together, they represent the key audiences for communicating biomedical research. The task is to generate summaries that faithfully reflect the abstract while aligning with the knowledge and needs of each target persona. We created the dataset in a two-step process as shown in Figure \ref{fig:PERCS_overview_hor}. First, we generated the summaries using LLMs. Then, we employed experts to ensure the faithfulness of the summaries, followed by human evaluation of summary quality to assess alignment with different personas. We describe both steps in detail below.


\begin{figure*}
    \centering
    \includegraphics[width=0.75\textwidth,height=0.3\textheight,keepaspectratio]{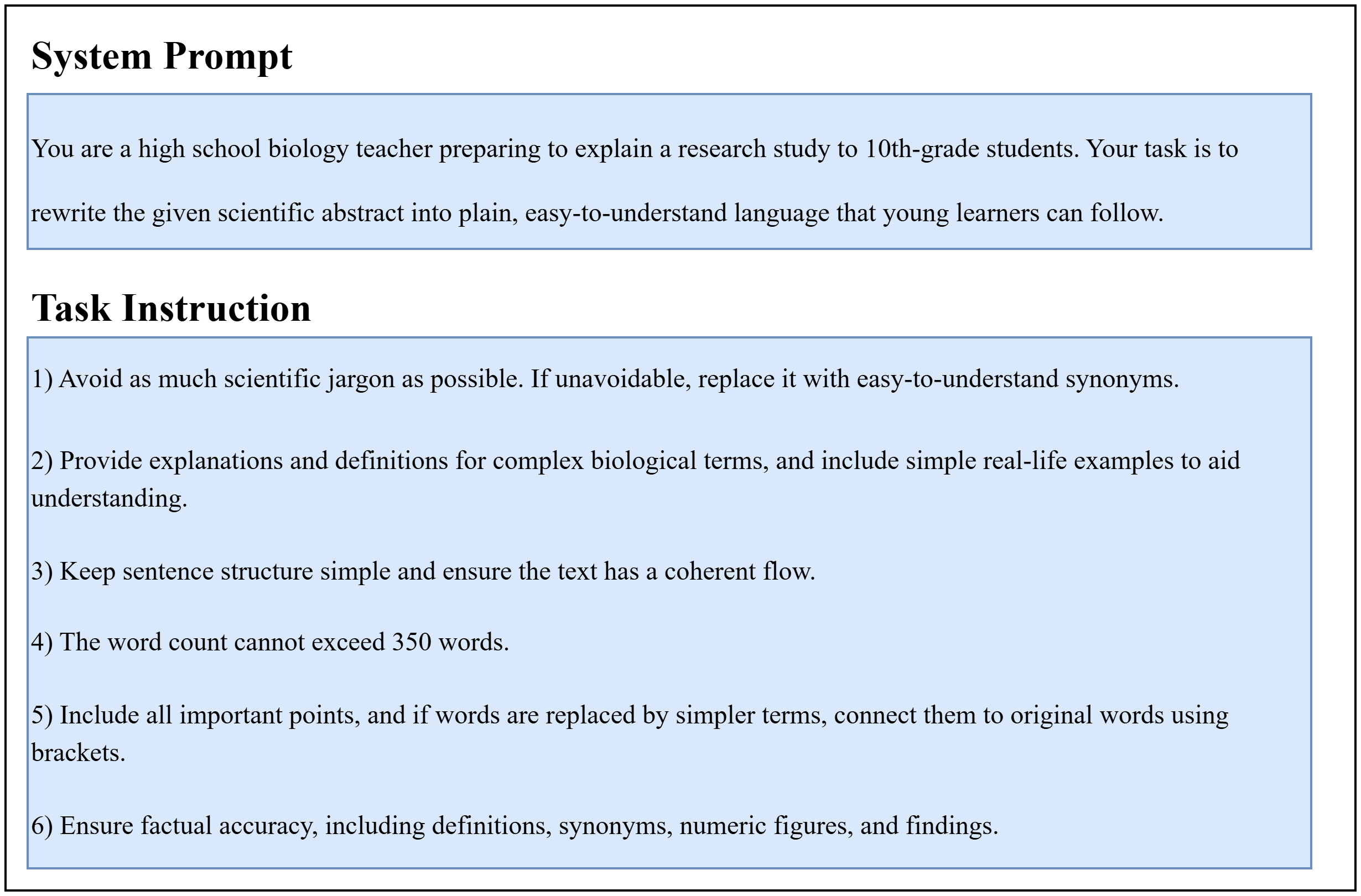}
    \caption{Example of a prompt design used for Lay persona summary generation in the PERCS dataset}
    \label{fig:lay_prompt}
\end{figure*}

\subsection*{Automatic Summary Generation}
To begin dataset creation, we selected biomedical abstracts to ensure coverage of diverse medical topics. We first collected 250 abstracts from \textit{PLOS Medicine}, \textit{PLOS Biology}, and \textit{PLOS Neglected Tropical Diseases} using the PLOS API. Queries targeted chronic diseases such as cancer, diabetes, Alzheimer’s disease, and chronic kidney disease, as well as common infectious conditions including typhoid and pneumonia. Articles were retrieved by searching for disease-related keywords in titles and abstracts and were manually reviewed for relevance.
To further expand topic coverage, we incorporated 250 additional abstracts from the PLABA dataset, a biomedical lay-summarization corpus that includes conditions such as COVID-19, kidney stones, and cystic fibrosis. Abstracts were randomly sampled from PLABA’s training split to avoid overlap with the PLOS subset. The final PERCS dataset thus contains 500 biomedical abstracts spanning a broad range of diseases and study designs.

Previous research has shown that few-shot prompting or fine-tuning of large language models, such as GPT-3.5, can yield high-quality lay summaries even with limited data \cite{turbitt-etal-2023-mdc,goldsack-etal-2024-overview}. More recent work demonstrates that modern models can also produce strong summaries in zero-shot settings \cite{goldsack2025leveraging,agarwal2025overview}. Studies using GPT-4 and Mistral-Large-Instruct-2407 have reported superior human-rated performance on biomedical summarization tasks compared with fine-tuned domain-specific models such as BioMistral \cite{salvi-etal-2025-towards}. Based on these findings, we selected GPT-4 to generate the initial persona-specific summaries in PERCS.
Each abstract was paired with four model-generated summaries, one for each persona, using customized prompts designed to match the background knowledge and information needs of the target reader. Figure~\ref{fig:lay_prompt} highlights the prompt structure used for the layperson persona, and all complete persona-specific prompts are provided in Table~\ref{tab:prompt-percs} in the Appendix.
 This automated stage produced high-quality drafts intended as starting points for expert refinement. We acknowledge that such summaries may contain factual inaccuracies or misaligned tone, and therefore, expert review and correction were applied to ensure faithfulness and appropriate persona alignment.

\subsection*{Medical Expert Annotation \& Evaluation of Summary Quality} While LLM-generated summaries are effective, prior research shows they remain prone to factual errors and hallucinations when adapted for public use \cite{fang-etal-2024-understanding}. To address these challenges, we examined prior work on error categorization in lay summaries \cite{guo-etal-2024-appls,joseph-etal-2024-factpico}, incorporated persona-specific considerations, and developed a taxonomy of error types (Table~\ref{tab:error-types}) to guide evaluation of summaries for factual accuracy and persona alignment. We defined 11 fine-grained error types grouped into four categories (Table~\ref{tab:error-types}):  
(1) \textit{New Information Errors} (incorrect definitions, synonyms, or background)  
(2) \textit{Inference Errors} (contradictions, omissions, misinterpretations, or entity errors)  
(3) \textit{Structure and Readability Errors}  
(4) \textit{Hallucinations}

\begin{figure*}[!ht]
    \centering
    \includegraphics[width=\textwidth]{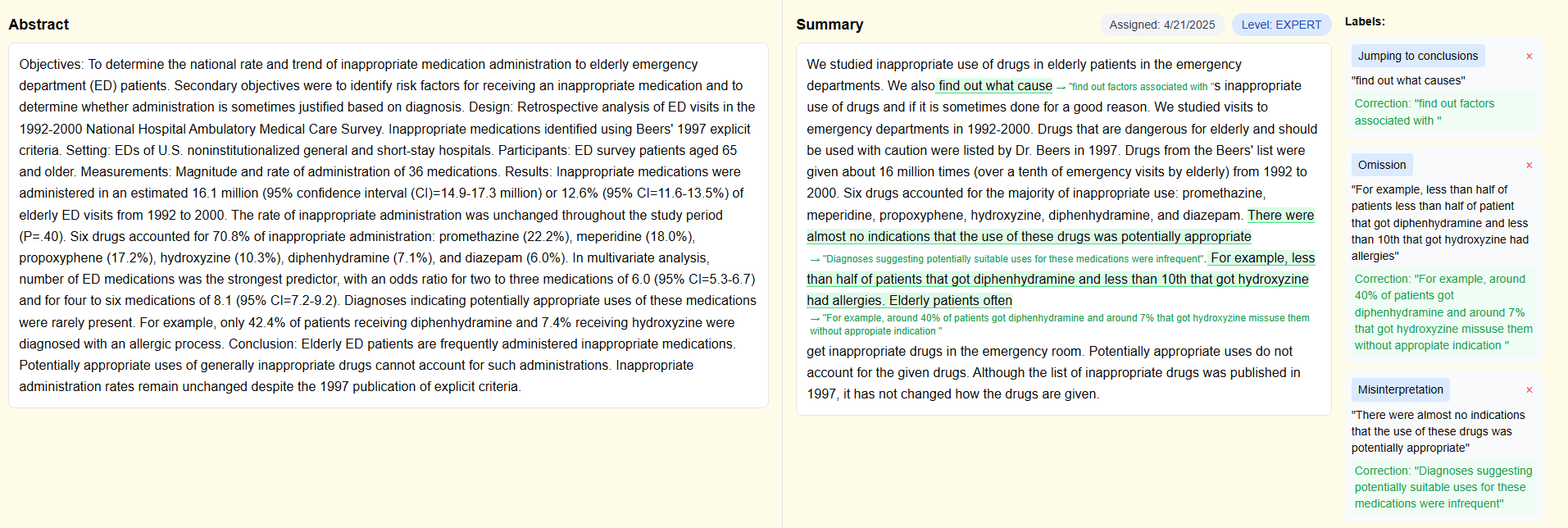}
    \caption{Example of addressing logical errors and supplying persona-specific missing information to produce an appropriate summary for the persona.}
    \label{fig:my-image}
\end{figure*}
\noindent\textbf{Step 1: Expert Annotation.} 
We employed two practicing physicians (MDs) to reviewed each summary for accuracy and revised it according to our taxonomy.

\begin{itemize}
    \item \textit{Annotation Task:} Using a web-based interface, the experts reviewed each summary alongside its source abstract and persona type. Based on predefined guidelines (provided in the Appendix), they highlighted erroneous segments, assigned error tags, and proposed corrections to improve persona relevance and faithfulness while maintaining textual coherence (example shown in Figure \ref{fig:PERCS_overview_hor}).

    \item \textit{Training:} Experts completed two rounds of calibration on 32 summaries (eight per persona) followed by reconciliation sessions to align interpretations of error labels and persona relevance. Consensus on labeling criteria was established before the main annotation phase.

    \item \textit{Inter-Annotator Agreement:} Given the open-ended nature of summarization, we assessed inter-annotator agreement by comparing sentence-level error labels (as shown in Figure \ref{fig:my-image}). On a random set of 40 summaries containing 420 sentences, we observed strong agreement, with Krippendorff’s alpha ranging from 0.986 to 0.998 across personas and 0.992 overall (see Table~\ref{tab:groupwise-agreement} for the complete agreement scores). After confirming agreement, the experts reviewed all 2,000 summaries for factual accuracy and persona alignment.
\end{itemize}

\begin{table}[ht]
\caption{Agreement between expert annotators for sentence-level error analysis (Krippendorff's Alpha)}
\centering
\footnotesize
\begin{tabular}{lcc}
\toprule
\textbf{Persona} & \textbf{Sentences} & \textbf{Krippendorff's Alpha} \\
\midrule
LAYMAN     & 137 & 0.995 \\
PREMED     & 108 & 0.997 \\
RESEARCHER & 106 & 0.998 \\
EXPERT     & 69  & 0.986 \\
OVERALL    & 420 & 0.992 \\
\bottomrule
\end{tabular}

\label{tab:groupwise-agreement}
\end{table}

\noindent\textbf{Step 2: Persona-Based Quality Evaluation.}  
Once the faithful summaries were finalized, we evaluated their quality to ensure proper alignment with the personas. Following prior work on lay-summary evaluation \cite{goldsack-etal-2023-enhancing,salvi-etal-2025-towards}, we used a five-point Likert scale with rubric-based guidance and two human raters per persona. Specifically, we adopted the rubric structure from Salvi et.al\cite{salvi-etal-2025-towards}, which assumes a target reader with high school-level biology knowledge. The rubric scores summaries on \emph{comprehensiveness}, \emph{layness}, and \emph{factuality} using a 1--5 scale:

\begin{itemize}
\item \textbf{Comprehensiveness:} Assesses whether essential topics, approaches, and findings are present (1 = incomplete; 3 = topic understood but key details missing; 5 = fully covered).
\item \textbf{Layness:} Measures how well jargon is reduced and concepts are explained using simple sentences, definitions, and analogies (1 = little simplification; 3 = mix of jargon and simple terms; 5 = plain, well-explained, easy to follow).
\item \textbf{Factuality:} Evaluates alignment with the abstract (1 = misrepresents findings or methods; 3 = mostly accurate with minor inconsistencies; 5 = fully faithful).
\end{itemize}
Because we covered diverse reader groups, we also added a \textbf{\emph{usefulness}} dimension, which evaluates whether the summary helps the intended reader grasp what was studied, why it matters, and what was found (1 = not helpful; 5 = very helpful). 

\textbf{Inter annotator agreement:} Each persona-specific reader evaluated 50 summaries tailored to their persona. Raters first read the abstract, then the summary, and recorded four ratings: \textit{comprehensiveness}, \textit{layness}, \textit{factuality}, and \textit{usefulness}. Overall, ratings were consistently high for comprehensiveness, faithfulness, and usefulness, averaging around 4.5 ± 0.3 across personas. Layness scores showed greater variation, as expected since the rubric was originally designed for lay summaries. Across personas, Krippendorff’s alpha values ranged from approximately 0.79 to 1.0, indicating strong inter-rater reliability overall. For full detailed results, including per-persona facet scores and agreement coefficients, see Table \ref{tab:persona_alpha}.

\begin{table*}
\caption{Human Evaluation Scores and Krippendorff's Alpha for Each Persona Across Four Criteria}
\label{tab:persona_alpha}
\centering
\footnotesize
\renewcommand{\arraystretch}{1.2}
\begin{tabular}{lcccc}
\hline
\textbf{Persona} & \textbf{Metric} & \textbf{Score} & \textbf{K-$\alpha$} \\
\hline
\multirow{4}{*}{Layman} 
    & Comprehensiveness & 4.94 &  0.82\\
    & Layness           & 4.98 & 1.0 \\
    & Faithfulness      & 5.0 & 1.0 \\
    & Usefulness        & 4.98 & 1.0 \\
\hline
\multirow{4}{*}{Pre-med} 
    & Comprehensiveness & 4.53 & 0.7933 \\
    & Layness           & 4.00 & 0.8792 \\
    & Faithfulness      & 4.34 & 0.8348 \\
    & Usefulness        & 4.13 & 0.8353 \\
\hline
\multirow{4}{*}{Researcher} 
    & Comprehensiveness & 4.52 & 0.9085 \\
    & Layness           & 4.41 & 0.9068 \\
    & Faithfulness      & 4.75 & 0.8997 \\
    & Usefulness        & 4.65 & 0.9086 \\
\hline
\multirow{4}{*}{Expert} 
    & Comprehensiveness & 4.47 & 0.989 \\
    & Layness           & 1.589 & 0.998 \\
    & Faithfulness      & 4.707 & 0.991 \\
    & Usefulness        &4.568 & 0.9986 \\
\hline
\end{tabular}
\end{table*}

\subsection*{Data Records}

The PERCS dataset is publicly available at the Open Science Framework repository at \url{https://doi.org/10.17605/OSF.IO/UMBJ7} and contains 500 biomedical research abstracts paired with 2,000 expert-curated persona-specific summaries. Each record corresponds to a single abstract and includes four summaries tailored to distinct reader personas: layperson, pre-medical student, researcher from a non-medical field, and medical expert.

\paragraph{Data structure.}
The dataset is released in JSON format, where each entry represents one abstract with all four persona-specific summaries. Each entry contains the following fields:
\begin{itemize}
    \item \texttt{id}: Unique identifier for the abstract. For PLABA abstracts, this corresponds to the PMID, and for PLOS abstracts, it is the PLOS-assigned identifier.
    \item \texttt{abstract}: Full text of the abstract.
    \item \texttt{summaries}: Object containing four expert-corrected summaries, keyed by persona:
    \begin{itemize}
        \item \texttt{layman}: Summary for general public readers.
        \item \texttt{premed}: Summary for pre-medical students.
        \item \texttt{researcher}: Summary for researchers from non-medical fields.
        \item \texttt{expert}: Summary for medical experts.
    \end{itemize}
\end{itemize}

\noindent An illustrative example of persona variation in tone, terminology, and level of detail is shown in Figure~\ref{fig:persona-summary-examples}.

\paragraph{Data splits}
To support benchmarking and reproducibility, the dataset is divided into two non-overlapping splits: 350 abstracts (70\%) for training and 150 abstracts (30\%) for testing. Both subsets maintain balanced representation across sources (PLOS and PLABA). No abstracts appear in both splits.

\paragraph{License}
The dataset is released under the Creative Commons Attribution 4.0 International (CC BY 4.0) license, allowing unrestricted reuse, with appropriate attribution.

\section*{Technical Validation}
\subsection*{Topic Distribution}  
The abstracts cover a wide range of diseases. PLOS sources focus on chronic conditions such as cancer, diabetes, and Alzheimer’s disease, while PLABA adds topics such as arthritis, kidney stones, and COVID-19. The complete topic distribution for PLOS abstracts is provided in Table~\ref{tab:topic-percs} in the Appendix.

\subsection*{Summary Characteristics}  
For each persona, we measured summary length (word count), lexical diversity using Type–Token Ratio (TTR) \cite{templin1957certain} and Measure of Textual Lexical Diversity (MTLD) \cite{mccarthy2005assessment}, readability using the Dale–Chall Readability Score (DCRS) \cite{chall1995readability}, and Coleman–Liau Index (CLI) \cite{coleman1975computer}, and semantic overlap with the abstract using BERTScore (BERT\_S) \cite{zhang2019bertscore}. We computed each metric for individual summaries and report their mean and variation in Table~\ref{tab:percs-stats}. We found that layperson summaries were longer, with added definitions and context, but showed lower lexical diversity and the least overlap with the abstract (BERTScore). They were also easier to read (lower DCRS, CLI). As expertise increased, summaries became shorter, more complex (higher DCRS, CLI), more diverse (higher TTR, MLTD), and closer to the abstract (higher BERT\_S). Pre-med and researcher summaries differed modestly, with researchers using slightly more technical phrasing and achieving marginally higher readability scores and abstract overlap. Expert summaries were the most concise, lexically rich, and information-dense.

\begin{table}[htbp]
\centering
\caption{PERCS Dataset Statistics: Average Readability, Word Count, and Lexical Diversity by Persona}
\label{tab:percs-stats}
\footnotesize
\begin{tabular}{l|cccccc}
\hline
\textbf{Persona} & \textbf{Word Count} & \textbf{DCRS} & \textbf{CLI} & \textbf{BERT\_S} & \textbf{TTR} & \textbf{MLD} \\
\hline 
LAYMAN      & 270.91 &  8.17 & 10.48 & 0.811 & 0.559 &  98.159 \\
PREMED      & 243.48 &  9.96 & 14.31 & 0.854 & 0.598 & 102.845 \\
RESEARCHER  & 246.36 & 10.12 & 14.58 & 0.865 & 0.599 & 102.779 \\
EXPERT      & 166.43 & 11.76 & 17.18 & 0.890 & 0.678 & 114.860 \\
\hline
\end{tabular}
\end{table}

\subsection*{Experimental Benchmarking}

We evaluated four large language models (LLMs), including GPT-4o, Mistral-8-7B-Instruct, Gemini-2.0 Flash Lite, and LLaMA-3 70B, on the PERCS dataset, which provides persona-specific summaries for four reader groups: laypersons, pre-medical students, non-medical researchers, and medical experts.

\subsubsection*{Experimental Setup}

We benchmarked three standard prompting strategies for LLM-based summarization:

\begin{itemize}
    \item \textbf{Zero-shot:} The model receives the persona-specific prompt used to create the dataset and generates a summary without additional examples.
    \item \textbf{Few-shot:} The same persona-specific prompt is provided along with three in-domain exemplars from the dataset.
    \item \textbf{Self-refine:} The model first produces a zero-shot summary, then critiques it for persona alignment and faithfulness to the abstract, provides self-feedback, and revises iteratively until satisfied. In this setup, the LLM acts as both feedback provider and evaluator.
\end{itemize}

We evaluated summaries on three key aspects: comprehensiveness, readability, and faithfulness, using automatic evaluation metrics. Comprehensiveness was measured using ROUGE-1, ROUGE-2, ROUGE-L \cite{lin2004rouge} and SARI \cite{xu2016optimizing}; readability using FKGL \cite{kincaid1975derivation,flesch2007flesch}, DCRS \cite{chall1995readability}, CLI \cite{coleman1975computer}, and LENS \cite{maddela2022lens}; and faithfulness using SummaC (Conv) \cite{laban2022summac}. For summary generation, we accessed GPT-4o via the official OpenAI API platform. Other open-source models, including LLaMA3, Mistral, and Gemini, were queried using the OpenRouter API interface. For automatic evaluation, we computed ROUGE, SARI, FKGL, DCRS, and CLI using publicly available Python libraries. LENS and SummaC metrics, which require more computational resources, were run on a machine equipped with a T4 GPU (16 GB VRAM).

\begin{table*}[]
\caption{Selected performance on PERCS across personas using automatic evaluation metrics. The readability metrics (FKGL, DCRS, CLI, and LENS) directionality is not defined in a persona-based setting.}
\label{tab:main_result}
\begin{center}
\footnotesize
\renewcommand{\arraystretch}{1.1}
\setlength{\tabcolsep}{4pt}
\begin{tabular}{p{1.5cm}|p{1.5cm}p{2.5cm}*{9}{c}}
\hline
\textbf{Persona} & \textbf{Model} & \textbf{Method} 
& \textbf{R-1~↑} & \textbf{R-2~↑} & \textbf{R-L~↑}
& \textbf{SARI~↑} & \textbf{FKGL} & \textbf{DCRS}
& \textbf{CLI} & \textbf{LENS} & \textbf{SC~↑} \\
\hline \hline

\multirow{3}{*}{Layperson}
 & Gemini & Zero-shot  & 0.6029 & 0.2624 & 0.3691 & 53.0180 & 10.04 & 8.71 & 11.66 & 52.5033 & 0.3031 \\

  & GPT-4o                          & Few-shot  & 0.6016 & 0.2476 & 0.3554 & 53.5134 & 11.22 & 8.94 & 11.99 & 75.0276 & 0.2987 \\

  & GPT-4o                             & Self-Refine & 0.5288 & 0.1727 & 0.2696 & 48.3749 & 11.24 & 9.47 & 12.84 & 71.6981 & 0.2553 \\

\hline

\multirow{3}{*}{Premed}
   &Mistral  & Zero-shot  & 0.5819 & 0.2629 & 0.3622 & 49.3190 & 16.08 & 11.80 & 17.53 & 50.3914 & 0.3498 \\

  &Llama-3.1                            & Few-shot  & 0.6149 & 0.2926 & 0.3823 & 51.9053 & 16.72 & 11.78 & 17.69 & 53.6619 & 0.2969 \\

  & Gemini                           & Self-Refine & 0.5564 & 0.2084 & 0.3011 & 46.4857 & 14.82 & 10.76 & 16.42 & 45.4320 & 0.2613 \\

\hline

\multirow{3}{*}{Researcher}
  &Llama-3.1 & Zero-shot & 0.6120 & 0.2880 & 0.3719 & 48.4270 & 15.19 & 10.02 & 14.95 & 57.7103 & 0.2769 \\

   &Mistral                             & Few-shot   & 0.6122 & 0.3072 & 0.3978 & 49.4988 & 15.05 & 10.65 & 15.93 & 59.3144 & 0.3928 \\

  &Gemini                              & Self-Refine  & 0.5564 & 0.2084 & 0.3011 & 46.6575 & 14.82 & 10.76 & 16.42 & 50.6804 & 0.2803 \\

\hline

\multirow{3}{*}{Expert}
  &GPT-4o & Zero-shot  & 0.6000 & 0.2905 & 0.3997 & 42.1944 & 17.91 & 12.77 & 19.44 & 51.2367 & 0.3283 \\

  &Gemini                             & Few-shot  & 0.6599 & 0.3815 & 0.4912 & 44.6463 & 15.43 & 12.57 & 18.03 & 54.1410 & 0.3795 \\

  &GPT-4o                             & Self-Refine & 0.5577 & 0.2547 & 0.3535 & 42.4489 & 17.30 & 12.76 & 19.39 & 51.3194 & 0.2946 \\

\hline 

\end{tabular}
\end{center}
\end{table*}

\subsection*{Results}
To enable direct comparison within each persona and evaluation facet, all metric scores were normalized using Min–Max scaling after adjusting the directionality of the readability metrics. For the layperson, pre-medical, and researcher personas, the readability metrics FKGL, DCRS, and CLI were inverted, since lower values indicate simpler and more accessible language, and were therefore treated as higher normalized scores. For the expert persona, the LENS metric was inverted instead, as lower raw LENS values correspond to more concise and domain-appropriate phrasing. This procedure ensured that all normalized metrics were directionally aligned such that higher values consistently indicated better performance, enabling comparison of model performance across comprehensiveness, readability, and faithfulness facets. Following normalization, all models and prompting methods were ranked within each persona. Table~\ref{tab:main_result} presents the best-performing model for each prompting strategy across the personas. The complete results for each persona are provided in Tables~\ref{tab:complete_results_premed}, \ref{tab:complete_results_researcher}, \ref{tab:complete_results_lay}, and \ref{tab:complete_results_expert} in the Appendix.

Across all personas, few-shot prompting generally achieved the best balance between comprehensiveness, readability, and faithfulness. The few-shot GPT-4o model obtained the highest overall performance for the lay summaries (R-1 = 0.602, SARI = 53.51), producing text that was both accurate and accessible. For pre-medical readers, few-shot LLaMA-3.1 performed best (R-1 = 0.615, R-L = 0.382, SARI = 51.91), indicating improved content coverage while maintaining moderate readability. Among researcher-level summaries, few-shot Mistral was able to provide the most precise and detailed information, achieving the highest faithfulness (SummaC = 0.393) and comprehensiveness (R-L = 0.398). Finally, for expert-level outputs, zero-shot GPT-4o yielded the best overall alignment between coverage and readability (R-L = 0.400, SC = 0.328), demonstrating that the LLM could effectively control generation for an expert persona without in-context examples.

Across prompting strategies, few-shot configurations consistently provided higher normalized composite scores, whereas self-refinement offered limited gains and occasionally reduced lexical overlap. Collectively, these results indicate that in-context examples substantially improve persona-aware summarization quality, with GPT-4o and LLaMA-3.1 exhibiting the most stable performance across reader types, Mistral excelling in factual precision, and Gemini producing strong expert-oriented summaries with competitive ROUGE scores.

\begin{table}[htbp]
\centering
\scriptsize
\renewcommand{\arraystretch}{1.2}
\begin{tabular}{|p{1,2cm}|p{15.5cm}|}
\hline
\textbf{Model} & \textbf{Generated Summary} \\
\hline
\textbf{Abstract} & 
Background: Antiepileptic drugs (AEDs) can be associated with neurotoxic side effects including cognitive dysfunction, a problem of considerable importance given the usual long-term course of treatment. Pregabalin is a relatively new AED widely used for the treatment of seizures and some types of chronic pain including fibromyalgia. We measured the cognitive effects of 12 weeks of pregabalin in healthy volunteers.[.......]. These cognitive effects were paralleled by complaints on the Portland Neurotoxicity Scale, a subjective measure of neurotoxicity (p < 0.01). Conclusion: At conventional doses and titration, pregabalin induced mild negative cognitive effects and neurotoxicity complaints in healthy volunteers. These effects are one factor to be considered in the selection and monitoring of chronic AED therapy. Class of Evidence: This study provides Class I evidence that pregabalin 300 mg BID negatively impacts cognition on some tasks in healthy volunteers. \\
\hline
\textbf{GPT-4o} & 
Antiepileptic drugs (AEDs) are medicines used to treat seizures and chronic pain like fibromyalgia. However, they can sometimes have negative effects on the brain, making it hard to think clearly or remember things. This is important because people usually need these medicines for a long time. Pregabalin is a newer AED that we wanted to test for these effects. [.......]. This means they experienced some negative changes in how their brains felt or worked (neurotoxicity), like feeling foggy or having trouble focusing. In conclusion, when pregabalin is taken at typical doses, it can mildly affect brain functions related to thinking and memory in healthy people. These effects are important for doctors to consider when prescribing pregabalin, especially for people who will take it long-term to manage chronic conditions. \\
\hline
\textbf{Mistral} & 
Antiepileptic drugs (AEDs), used for treating seizures and certain types of chronic pain, can sometimes cause side effects like problems with thinking and memory. These issues are important to consider since AEDs are often taken for long periods. Pregabalin is a relatively new AED that is widely used for seizures and pain management, including in a condition called fibromyalgia. [.......]. Moreover, the volunteers who received pregabalin also reported more complaints on the Portland Neurotoxicity Scale, a questionnaire used to assess neurotoxicity symptoms, meaning that they noticed more issues with their thinking and memory (p < 0.01). In conclusion, taking pregabalin at conventional doses and titration for 12 weeks caused mild negative cognitive effects and neurotoxicity complaints in healthy volunteers. This information is essential when selecting and monitoring long-term AED therapy. The study provides Class I evidence that pregabalin 300 mg twice daily negatively impacts cognition on some tasks in healthy volunteers. \\
\hline
\textbf{Gemini} & 
This research looked at how a drug called pregabalin affects the brain. Pregabalin is used to treat epilepsy (seizures) and some types of long-term pain. However, some medications for epilepsy can sometimes cause problems with thinking and memory. These are known as neurotoxic side effects. [.......]. It did not have a positive effect for the participants. In addition, those taking pregabalin reported more common side effects related to neurotoxicity, based on their answers to the questionnaire. This suggests that the drug had some unwanted effects on brain function and mental awareness. Based on the results of this research, the study concluded that pregabalin, used at a typical everyday dose, caused some negative outcomes in cognition and signs of neurotoxicity among the subjects. These observations are important to be taken into account when doctors choose and monitor long-term treatments using this type of medicine. The results provide dependable evidence that pregabalin at a dose of 300 mg twice a day negatively affects thinking in healthy individuals. \\
\hline
\textbf{Llama 3.1} & 
Pregabalin is a relatively new medicine used to treat seizures and certain types of chronic pain, including a condition called fibromyalgia. However, like many medicines that affect the brain, it can have side effects on cognitive function, which includes things like memory, attention, and problem-solving skills. This is a significant concern because people often take these medicines for a long time. [.......]. Additionally, people taking pregabalin reported more complaints related to neurotoxicity, which includes symptoms like feeling tired, having trouble concentrating, or experiencing memory problems. These findings suggest that pregabalin, at the usual prescribed dose, can cause mild but noticeable negative effects on cognitive function in healthy individuals. This information is important for doctors to consider when prescribing pregabalin for long-term treatment, as they need to weigh the benefits against the potential risks to cognitive health. \\
\hline
\end{tabular}
\caption{Example summaries for Lay persona generated by the LLMs in a few-shot setting.}
\label{tab:gen_sum_ex_methods_hidden}
\end{table}

\subsection*{Discussion}

The results demonstrate that the prompting strategy has a substantial impact on persona-aligned summarization performance. Few-shot prompting consistently achieved the best trade-off between coverage, readability, and faithfulness across all reader personas. This finding suggests that even a small number of in-context examples can effectively guide large language models to adjust lexical choice, syntactic complexity, and informational detail to match the intended audience persona. By contrast, zero-shot prompting produced strong but less consistent results, indicating that LLMs contain considerable generalization capacity for persona-based summarization without explicit fine-tuning. Self-refinement provided smaller improvements, highlighting limited gains from self-critique.

GPT-4o and LLaMA-3.1 showed the most stable performance across all personas. They were able to balance information detail and clarity while maintaining the readability level desired for each persona. Both models achieved this by using clear and easy-to-follow language and including technical or scientific terms only when necessary, which preserved the main ideas without overwhelming the reader. For the lay summaries, such as the example shown in Table \ref{tab:gen_sum_ex_methods_hidden}, GPT-4o produced fluent and easy-to-follow sentences that explained the main findings in simple terms while preserving the original meaning. LLaMA-3.1 also used appropriate information and simplification, ensuring summaries were appropriate for the persona, which was reflected in its consistently good readability and comprehensiveness metric results.

Gemini's summaries, on the other hand, were often too long. They tended to include excessive background information or repeat details from the abstract. This made the summaries difficult to follow, particularly for lay readers. In the example shown in Table \ref{tab:gen_sum_ex_methods_hidden}, Gemini clearly produced the longest summary. It also used technical words such as "neurotoxicity," which GPT-4o and LLaMA-3.1 avoided by using simpler phrases to describe the same concept. While Gemini's summaries were comprehensive, this approach reduced clarity and made the text less suitable for readers without a technical background. Mistral took the opposite approach. It created short and simple summaries that were easy to read. However, Mistral often extracted relevant but unnecessary details for the persona from the abstract, such as the mention of a p-value, which is not useful for lay readers. It also often omitted definitions or brief explanations of key terms, which could make it harder for lay readers or pre-med students to fully understand some biomedical concepts.

While few-shot prompting remains the most reliable and generalizable strategy, future work should explore hybrid methods that combine in-context examples with targeted feedback mechanisms to further improve faithfulness and information control in persona-based summarization. The PERCS dataset, therefore, serves as a comprehensive testbed for evaluating persona-guided, controllable biomedical summarization and for developing methods that balance readability, comprehensiveness, and factuality across diverse audiences.

\section*{Data Availability}
The dataset presented in this study is openly available in the Open Science Framework repository at \url{https://doi.org/10.17605/OSF.IO/UMBJ7}

\section*{Code Availability}
All code required to reproduce the dataset and the benchmarking experiments reported in this study is available on GitHub at \url{https://github.com/rohancsalvi/PERCS-Dataset}.

\bibliography{sample}

\section*{Appendix}

\subsection*{Evaluation Guidelines for Expert Error-Based Summary Annotation}
\label{appendix:evaluation-guidelines}

\textbf{Objective:} \\
Evaluate model-generated plain language summaries of biomedical research abstracts. Annotators assess each summary for accuracy, completeness, readability, and reader appropriateness, and label specific errors based on the taxonomy below.\\

\textbf{Materials Provided:}
\begin{itemize}
    \item Original abstract of the medical article
    \item Model-generated plain language summary
\end{itemize}

\textbf{Evaluation Procedure:}
\begin{enumerate}
    \item \textbf{Review the Abstract:} Read the abstract carefully to understand its methodology, key findings, and context.
    \item \textbf{Evaluate the Summary:} Compare the summary to the abstract and your medical expertise.
    \item \textbf{Annotate Errors:} Highlight problematic segments in the summary, tag each with the corresponding error type, and suggest corrections when necessary.
\end{enumerate}

\vspace{1em}

\textbf{Error Categories:}

\textbf{1. New Information Errors (Expert Judgment Required)}
\begin{itemize}
    \item \textit{Incorrect Definitions:} Inaccurate or misleading explanations of medical terms or concepts.
    \item \textit{Incorrect Synonyms:} Inappropriate simplification or substitution that changes medical meaning.
    \item \textit{Incorrect Background Information:} Medically inaccurate or misleading added context.
\end{itemize}

\textbf{2. Inference Errors (Check Against Abstract)}
\begin{itemize}
    \item \textit{Contradiction:} Information directly conflicts with the abstract.
    \item \textit{Omission:} Important findings or details from the abstract are missing.
    \item \textit{Jumping to Conclusions:} Unsupported generalizations or causal claims.
    \item \textit{Misinterpretation:} Inaccurate paraphrasing or simplification of abstract content.
    \item \textit{Entity Errors:} Factual errors in names, numbers, dosages, or statistics.
\end{itemize}

\textbf{3. Structure and Readability Errors}
\begin{itemize}
    \item \textit{Structure Error:} Poor organization or illogical flow.
    \item \textit{Grammatical Error:} Issues in spelling, punctuation, or sentence structure.
    \item \textit{Persona Relevance:} Language is too technical or too simplistic for the target reader.
\end{itemize}

\textbf{4. Hallucinations}
\begin{itemize}
    \item \textit{Hallucination:} Fabricated, irrelevant, or nonsensical content not supported by the abstract or general medical knowledge.
\end{itemize}

\textbf{Annotation Instructions:}
\begin{itemize}
    \item Highlight the erroneous segment in the summary.
    \item Tag the segment with the appropriate error label (e.g., \texttt{Contradiction}, \texttt{Incorrect Definition}).
    \item Provide a corrected version using either the abstract or medical expertise.
\end{itemize}

\begin{table*}[b]
\centering
\scriptsize

\renewcommand{\arraystretch}{1.2}
\caption{Topic Distribution of PLOS abstracts in PERCS: Test Set (75 abstracts) and Entire Dataset (250 abstracts)}
\label{tab:topic-percs}
\begin{tabular}{|l|c|c|}
\hline
\textbf{Disease} & \textbf{Test} & \textbf{Entire} \\
\hline
Cancer                            & 10 & 25  \\
Diabetes                          & 10 & 25  \\
Tuberculosis                      & 10 & 25  \\
Hepatitis                         & 10 & 25  \\
Chronic Kidney Disease            & 5  & 25  \\
HIV/ AIDS                         & 5  & 25  \\
Cardiovascular Diseases           & 5  & 20  \\
Obesity                           & 4  & 15  \\
Alzheimer's                       & 4  & 15  \\
Hypertension                      & 3  & 10  \\
Stroke                            & 3  & 10  \\
Pneumonia                         & 2  & 10  \\
Malaria                           & 2  & 10  \\
Salmonella/Typhoid               & 2  & 10  \\
\hline
\textbf{Total}                    & \textbf{75} & \textbf{250} \\
\hline
\end{tabular}
\end{table*}

\begin{table*}
\centering
\scriptsize
\caption{Initial summary generation prompts for each persona used in the creation of the PERCS dataset.}
\label{tab:prompt-percs}
\begin{tabular}{p{0.08\textwidth} p{0.8\textwidth}}
\hline
\textbf{Persona} & \textbf{Prompt} \\ \hline

\textbf{Expert} &
You are a seasoned subject matter expert in the biological sciences and are preparing a summary of a research abstract for an expert audience in the same field. You decide to generate this summary with the following key principles in mind:\par
1) The target audience is experts and professionals within the same scientific field who are seeking a concise overview of the study.\par
2) Do not simplify technical terms or scientific concepts; preserve their complexity and meaning.\par
3) Maintain a formal and objective tone suitable for a professional scientific discourse.\par
4) The sentence structure should be academically appropriate and support a coherent, logical flow of information.\par
5) Ensure the summary includes the core findings, methodology, and significance of the study without omitting critical data.\par
6) The text must remain factually accurate, including all relevant terminology, experimental parameters, numeric results, and implications.\par
The word count should not exceed 250 words.\\[12pt]

\textbf{Researcher} &
You are a knowledgeable science communicator with interdisciplinary expertise. Your task is to write a clear and comprehensive summary of a scientific abstract for researchers outside the specific field of the study but with a solid understanding of general scientific principles. When summarizing, follow these key principles:\par
1) The audience consists of researchers in other fields than biology and medicine who may not be familiar with domain-specific terminology or methods.\par
2) Use clear and accessible language. Avoid unnecessary technical jargon; when technical terms are necessary, explain them briefly in context.\par
3) Accurately represent the study's main findings, methodology, and results.\par
4) Maintain a professional but approachable tone. Aim for clarity and precision without overly simple or domain-specific language.\par
5) Structure the summary as a paragraph. Do not use bullet points, numbered lists, or Q\&A formats.\par
6) Focus on the scientific content only. Do not include commentary about the summarization process or subjective judgement of the study's importance.\par
The final summary should not exceed 350 words.\\[12pt]

\textbf{Pre-Med} &
You are an experienced academic tutor in the medical sciences. Your task is to write a summary of a research abstract specifically for pre-med students. When summarizing, follow these key principles:\par
1) The audience has foundational knowledge of biology and chemistry.\par
2) Use clear, simple language, and avoid technical jargon unless essential; when used, briefly explain the term.\par
3) Maintain accuracy while making the information accessible and engaging for pre-med students.\par
4) Ensure the summary flows logically and reads like a helpful explanation, not as dense scientific writing.\par
5) Ensure the summary accurately conveys the study’s key findings, methodology, and relevance to medical sciences.\par
6) The summary should be written as a continuous paragraph avoiding bullet points, lists, or Q\&A formats.\par
The word count of the summary should not exceed 350 words. Do not include any concluding commentary for the students, notes on the summarization process, or any self-determined importance of the study for pre-med students.\\[12pt]

\textbf{Layman} &
You are a biology teacher in a high school and want to teach students in 10th grade about a research study. Your goal is to convey the information in the abstract in plain and easy-to-understand language that students can follow. You decide to generate a plain text for the same abstract keeping in mind what makes a text simple and easy to understand:\par
1) Avoid as much scientific jargon as possible. If unavoidable, replace it with easy-to-understand synonyms.\par
2) Provide explanations and definitions for complex biological terms, and include simple real-life examples to aid understanding.\par
3) Keep sentence structure simple and ensure the text has a coherent flow.\par
4) The word count cannot exceed 350 words.\par
5) Include all important points, and if words are replaced by simpler terms, connect them to original words using brackets.\par
6) Ensure factual accuracy, including definitions, synonyms, numeric figures, and findings.

The final summary should not exceed 350 words.
\\

\hline
\end{tabular}
\end{table*}

\begin{table*}[t]

\caption{Performance of Benchmarking Approaches for the Pre-med Persona on the PERCS Dataset Using Automatic Evaluation Metrics}
\label{tab:complete_results_premed}
\begin{center}
\scriptsize
\renewcommand{\arraystretch}{1.2}
\begin{tabular}{|p{1.3cm}|p{1cm}|*{11}{c|}}
\hline
\textbf{Method} & \textbf{Model}
& \textbf{R-1~↑} & \textbf{R-2~↑} & \textbf{R-L~↑}
& \textbf{SARI~↑} & \textbf{FKGL} & \textbf{DCRS}
& \textbf{CLI} & \textbf{LENS} & \textbf{SC~↑} \\
\hline

\multirow{4}{*}{Zero-shot}
      & GPT-4    & 0.5916 & 0.2380 & 0.3434 & 49.5828 & 17.91 & 12.77 & 19.44 & 46.2652 & 0.2904 \\
      & Gemini   & 0.5960 & 0.2495 & 0.3424 & 48.3962 & 15.47 & 12.69 & 18.34 & 49.8001 & 0.2772 \\
      & Mistral  & 0.5819 & 0.2629 & 0.3622 & 49.3190 & 16.08 & 11.80 & 17.53 & 50.3914 & 0.3498 \\
      & LLaMA 3  & 0.5900 & 0.2568 & 0.3429 & 49.0921 & 17.32 & 12.28 & 18.21 & 49.5902 & 0.2759 \\
\hline

\multirow{4}{*}{Few-shot}
      & GPT-4    & 0.6089 & 0.2550 & 0.3629 & 50.4423 & 15.74  & 11.12  & 16.73  & 45.8277 & 0.3014 \\
      & Gemini   & 0.6024 & 0.2683 & 0.3648 & 48.3962 & 15.43 & 12.57 & 18.03 & 53.2864 & 0.3006 \\
      & Mistral  & 0.5911 & 0.2686 & 0.3688 & 49.2427 & 16.04 & 11.80 & 17.46 & 48.1884 & 0.3593 \\
      & LLaMA 3  & 0.6149 & 0.2926 & 0.3823 & 51.9053 & 16.72 & 11.78 & 17.69 & 53.6619 & 0.2969 \\
\hline

\multirow{4}{*}{Self-Refine}
      & GPT-4    & 0.5624 & 0.2100 & 0.3126 & 47.8304 & 15.83 & 11.43 & 17.59 & 47.0772 & 0.2684 \\
      & Gemini   & 0.5564 & 0.2084 & 0.3011 & 46.4857 & 14.82 & 10.76 & 16.42 & 45.4320 & 0.2613 \\
      & Mistral  & 0.4980 & 0.2028 & 0.2800 & 46.6206 & 15.42 & 10.52 & 17.01 & 45.9425 & 0.3249 \\
      & LLaMA 3  & 0.4973 & 0.2073 & 0.2748 & 48.7969 & 16.42 & 10.07 & 15.98 & 46.0638 & 0.2568 \\
\hline

\end{tabular}
\end{center}
\end{table*}

\begin{table*}[t]
\caption{Performance of Benchmarking Approaches for the Researcher Persona on the PERCS Dataset Using Automatic Evaluation Metrics}
\label{tab:complete_results_researcher}
\begin{center}
\scriptsize
\renewcommand{\arraystretch}{1.2}
\begin{tabular}{|p{1.3cm}|p{1cm}|*{9}{c|}}
\hline
\textbf{Method} & \textbf{Model}
& \textbf{R-1~↑} & \textbf{R-2~↑} & \textbf{R-L~↑}
& \textbf{SARI~↑} & \textbf{FKGL} & \textbf{DCRS}
& \textbf{CLI} & \textbf{LENS} & \textbf{SC~↑} \\
\hline

\multirow{4}{*}{Zero-shot}
      & GPT-4    & 0.6207 & 0.2742 & 0.3831 & 43.9918 & 15.87 & 11.00 & 16.48 & 51.2367 & 0.2536 \\
      & Gemini   & 0.6093 & 0.2838 & 0.3881 & 43.1912 & 15.53 & 11.19 & 16.88 & 53.5571 & 0.2735 \\
      & Mistral  & 0.6165 & 0.3039 & 0.4086 & 49.5102 & 15.34 & 10.79 & 16.39 & 55.6439 & 0.3227 \\
      & LLaMA 3  & 0.6120 & 0.2880 & 0.3719 & 48.4270 & 15.19 & 10.02 & 14.95 & 57.7103 & 0.2769 \\
\hline

\multirow{4}{*}{Few-shot}
      & GPT-4    & 0.6259 & 0.2778 & 0.3878 & 49.1682 & 16.14 & 11.10 & 16.86 & 59.1560 & 0.3074 \\
      & Gemini   & 0.6242 & 0.2956 & 0.4000 & 44.9244 & 15.28 & 11.01 & 16.42 & 54.1410 & 0.2769 \\
      & Mistral  & 0.6122 & 0.3072 & 0.3978 & 49.4988 & 15.05 & 10.65 & 15.93 & 59.3144 & 0.3928 \\
      & LLaMA 3  & 0.6164 & 0.2987 & 0.3838 & 48.2722 & 15.40 & 10.14 & 15.23 & 60.6989 & 0.2710 \\
\hline

\multirow{4}{*}{Self-Refine}
      & GPT-4    & 0.5624 & 0.2100 & 0.3126 & 46.3540 & 15.83 & 11.43 & 17.59 & 51.3194 & 0.2815 \\
      & Gemini   & 0.5564 & 0.2084 & 0.3011 & 46.6575 & 14.82 & 10.76 & 16.42 & 50.6804 & 0.2803 \\
      & Mistral  & 0.4980 & 0.2028 & 0.2800 & 46.6433 & 15.42 & 10.52 & 17.01 & 51.2360 & 0.3638 \\
      & LLaMA 3  & 0.4973 & 0.2073 & 0.2748 & 48.1174 & 16.42 & 10.07 & 15.98 & 49.0160 & 0.2651 \\
\hline

\end{tabular}
\end{center}
\end{table*}

\begin{table*}[t]
\caption{Performance of Benchmarking Approaches for the Lay person Persona on the PERCS Dataset Using Automatic Evaluation Metrics}
\label{tab:complete_results_lay}
\begin{center}
\scriptsize
\renewcommand{\arraystretch}{1.2}
\begin{tabular}{|p{1.3cm}|p{1cm}|*{9}{c|}}
\hline
\textbf{Method} & \textbf{Model}
& \textbf{R-1~↑} & \textbf{R-2~↑} & \textbf{R-L~↑}
& \textbf{SARI~↑} & \textbf{FKGL} & \textbf{DCRS}
& \textbf{CLI} & \textbf{LENS} & \textbf{SC~↑} \\
\hline

\multirow{4}{*}{Zero-shot}
      & GPT-4    & 0.5972 & 0.2439 & 0.3390 & 53.0256 & 10.37 & 8.54 & 11.30 & 51.7573 & 0.2962 \\
      & Gemini   & 0.6029 & 0.2624 & 0.3691 & 53.0180 & 10.04 & 8.71 & 11.66 & 52.5033 & 0.3031 \\
      & Mistral  & 0.5235 & 0.2068 & 0.3186 & 48.3885 & 12.38 & 9.37 & 12.93 & 62.7845 & 0.3626 \\
      & LLaMA 3  & 0.5900 & 0.2585 & 0.3527 & 52.0627 & 12.07 & 8.59 & 11.86 & 53.1642 & 0.2815 \\
\hline

\multirow{4}{*}{Few-shot}
      & GPT-4    & 0.6016 & 0.2476 & 0.3554 & 53.5134 & 11.22 & 8.94 & 11.99 & 75.0276 & 0.2987 \\
      & Gemini   & 0.6018 & 0.2611 & 0.3773 & 53.2864 & 11.00 & 8.97 & 12.17 & 69.1026 & 0.3115 \\
      & Mistral  & 0.5073 & 0.1892 & 0.3014 & 47.2191 & 13.08 & 9.81 & 13.84 & 64.7531 & 0.3571 \\
      & LLaMA 3  & 0.5841 & 0.2542 & 0.3499 & 53.6619 & 12.06 & 8.70 & 11.99 & 78.4790 & 0.2882 \\
\hline

\multirow{4}{*}{Self-Refine}
      & GPT-4    & 0.5288 & 0.1727 & 0.2696 & 48.3749 & 11.24 & 9.47 & 12.84 & 71.6981 & 0.2553 \\
      & Gemini   & 0.5422 & 0.1874 & 0.2756 & 47.9891 & 9.44  & 8.45 & 11.14 & 71.5003 & 0.2616 \\
      & Mistral  & 0.4750 & 0.1609 & 0.2523 & 47.1213 & 12.49 & 9.55 & 13.71 & 57.7433 & 0.2868 \\
      & LLaMA 3  & 0.4233 & 0.1366 & 0.2008 & 47.4832 & 11.84 & 8.32 & 12.07 & 62.4744 & 0.2433 \\
\hline

\end{tabular}
\end{center}
\end{table*}

\begin{table*}[t]
\caption{Performance of Benchmarking Approaches for the Expert Persona on the PERCS Dataset Using Automatic Evaluation Metrics}
\label{tab:complete_results_expert}
\begin{center}
\scriptsize
\renewcommand{\arraystretch}{1.2}
\begin{tabular}{|p{1.3cm}|p{1cm}|*{9}{c|}}
\hline
\textbf{Method} & \textbf{Model}
& \textbf{R-1~↑} & \textbf{R-2~↑} & \textbf{R-L~↑}
& \textbf{SARI~↑} & \textbf{FKGL} & \textbf{DCRS}
& \textbf{CLI} & \textbf{LENS} & \textbf{SC~↑} \\
\hline

\multirow{4}{*}{Zero-shot}
      & GPT-4    & 0.6000 & 0.2905 & 0.3997 & 42.1944 & 17.91 & 12.77 & 19.44 & 51.2367 & 0.3283 \\
      & Gemini   & 0.6348 & 0.3451 & 0.4511 & 45.1591 & 15.47 & 12.69 & 18.34 & 53.5571 & 0.4193 \\
      & Mistral  & 0.6478 & 0.3911 & 0.4758 & 48.2700 & 16.08 & 11.80 & 17.53 & 55.6439 & 0.4688 \\
      & LLaMA 3  & 0.6594 & 0.3928 & 0.4738 & 47.8625 & 17.32 & 12.28 & 18.21 & 57.7103 & 0.3586 \\
\hline

\multirow{4}{*}{Few-shot}
      & GPT-4    & 0.5418 & 0.2416 & 0.3555 & 41.4523 & 15.57 & 11.39 & 16.84 & 59.1560 & 0.3459 \\
      & Gemini   & 0.6599 & 0.3815 & 0.4912 & 44.6463 & 15.43 & 12.57 & 18.03 & 54.1410 & 0.3795 \\
      & Mistral  & 0.6625 & 0.4035 & 0.4946 & 48.4338 & 16.04 & 11.80 & 17.46 & 59.3144 & 0.4878 \\
      & LLaMA 3  & 0.6769 & 0.4198 & 0.5037 & 45.4880 & 16.72 & 11.78 & 17.69 & 60.6989 & 0.3068 \\
\hline

\multirow{4}{*}{Self-Refine}
      & GPT-4    & 0.5577 & 0.2547 & 0.3535 & 42.4489 & 17.30 & 12.76 & 19.39 & 51.3194 & 0.2946 \\
      & Gemini   & 0.6010 & 0.3126 & 0.4184 & 44.1335 & 15.36 & 12.49 & 17.91 & 50.6804 & 0.3397 \\
      & Mistral  & 0.6552 & 0.3973 & 0.4852 & 48.3519 & 16.06 & 11.80 & 17.50 & 57.4792 & 0.4783 \\
      & LLaMA 3  & 0.4732 & 0.2345 & 0.2981 & 43.1135 & 17.33 & 11.95 & 18.10 & 49.0160 & 0.2549 \\
\hline

\end{tabular}
\end{center}
\end{table*}

\end{document}